\documentclass[10pt,twocolumn,letterpaper]{article}

\usepackage{cvpr}
\usepackage{times}
\usepackage{epsfig}
\usepackage{graphicx}
\usepackage{amsmath}
\usepackage{amssymb}
\usepackage{mathtools}

\DeclarePairedDelimiter{\floor}{\lfloor}{\rfloor}

\usepackage{fancyhdr}

\usepackage[listofformat=parens, justification=centering, subrefformat=subparens]{subfig}

\usepackage[pagebackref=true,breaklinks=true,letterpaper=true,colorlinks,bookmarks=false]{hyperref}

\usepackage{todonotes}

\cvprfinalcopy 

\def\cvprPaperID{34} 

\graphicspath{{./figures/}{./code/}}

\renewcommand\cap[3]{\caption[#2]{\label{#1}\textsc{#2}. \small\textit{#3}}}

\renewcommand\etal[1]{\textit{et al.}~\cite{#1}}
\renewcommand\sec[1]{Sec.~\ref{sec:#1}}
\newcommand\fig[1]{Fig.~\ref{fig:#1}}
\newcommand\sfig[1]{Fig.~\subref{fig:#1}}

\ifcvprfinal\fi
\begin{document}

\title{Toward Open-Set Face Recognition}

\author{Manuel G\"unther \and Steve Cruz \and Ethan M. Rudd \and Terrance E. Boult\\
Vision and Security Technology Lab, University of Colorado Colorado Springs\\
{\tt\small \{mgunther,scruz,erudd,tboult\}@vast.uccs.edu}
}

\maketitle
\thispagestyle{empty}

{
  \chead{\footnotesize This is a pre-print of the original paper accepted for publication in the CVPR 2017 Biometrics Workshop.}
  \lhead{}
  \thispagestyle{fancy}
}

\begin{abstract}
Much research has been conducted on both face identification and face verification, with greater focus on the latter.
Research on face identification has mostly focused on using closed-set protocols, which assume that all probe images used in evaluation contain identities of subjects that are enrolled in the gallery.
Real systems, however, where only a fraction of probe sample identities are enrolled in the gallery, cannot make this closed-set assumption.
Instead, they must assume an open set of probe samples and be able to reject/ignore those that correspond to unknown identities.
In this paper, we address the widespread misconception that thresholding verification-like scores is a good way to solve the open-set face identification problem, by formulating an open-set face identification protocol and evaluating different strategies for assessing similarity.
Our open-set identification protocol is based on the canonical labeled faces in the wild (LFW) dataset.
Additionally to the known identities, we introduce the concepts of known unknowns (known, but uninteresting persons) and unknown unknowns (people never seen before) to the biometric community.
We compare three algorithms for assessing similarity in a deep feature space under an open-set protocol: thresholded verification-like scores, linear discriminant analysis (LDA) scores, and an extreme value machine (EVM) probabilities.
Our findings suggest that thresholding EVM probabilities, which are open-set by design, outperforms thresholding verification-like scores.
\end{abstract}

\section{Introduction}
\label{sec:introduction}

Face recognition algorithms have been widely researched over the past decades, resulting in tremendous performance improvements, particularly over the past few years.
Even traditional face recognition algorithms, i.e., before the widespread use of deep networks, performed quite well on frontal images under good illumination \cite{gunther2016frice}, making them commercially viable for certain applications.
For instance, verification scenarios such as automated border control stations \cite{sanches2016automated} allow reasonable control of imaging conditions and subjects usually cooperate with the system -- those that do not are easily spotted by airport security personnel.
As of 2006, O'Toole \etal{otoole2007face} demonstrated that algorithmic solutions were able to outperform humans for such constrained recognition tasks.
Therefore, researchers have shifted their focus to more difficult conditions.

\begin{figure}[t!]
  \centering
  \includegraphics[width=.95\columnwidth]{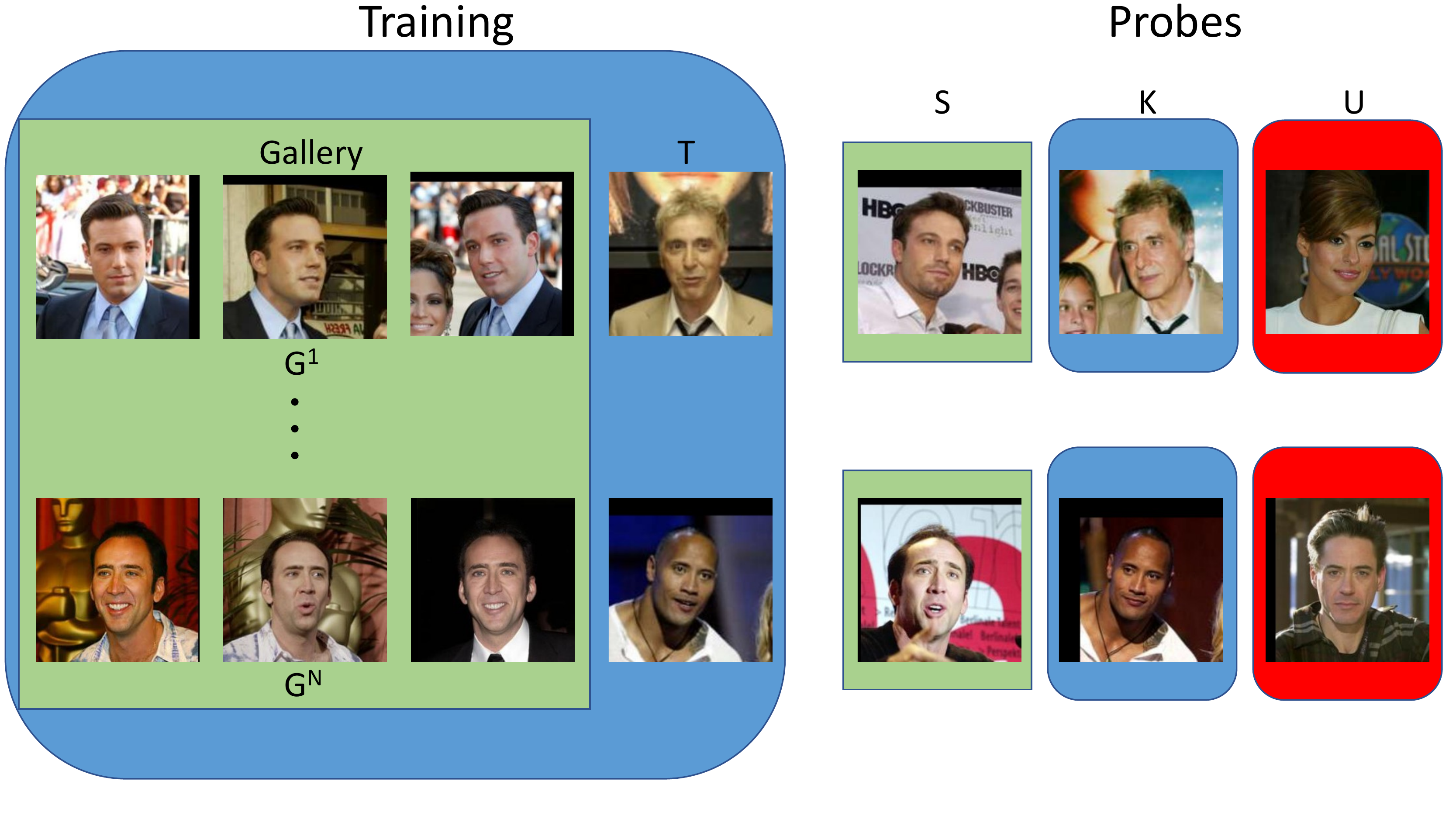}\\[-2ex]
  \cap{fig:teaser}{Open-Set Recognition}{Closed-set identification performs comparisons between the Gallery and known probe (S)ubjects.
The open-set identification protocol presented in this paper requires additional more subtle comparisons, due to the presence of known unknowns (uninteresting subjects) in the (T)raining set, (K)nown unknown probes of the same identity at query time, and (U)nknown unknowns whose identities are only seen during query time.
The open-set identification objective is to correctly identify probe (S)ubjects that are present in the gallery while rejecting all other probe queries as unknown.}
\end{figure}

G\"unther \etal{gunther2016frice} have found that these traditional face recognition algorithms are not designed to and, therefore, do not perform very well on images with uncontrolled factors such as facial expression, non-frontal illumination, partial occlusions of the face, or non-frontal face pose, which occur in modern face recognition datasets \cite{huang2007lfw,klare2015ijba}.
While different strategies have been proposed to improve the performance of traditional algorithms across pose, e.g., using face frontalization techniques \cite{hassner2015effective} or 3D modeling \cite{huq2007survey}, the introduction of deep convolutional neural networks (DCNNs) for face recognition \cite{taigman2014deepface,parkhi2015deep} has overcome the pose issue to a significant extent.
For example, deep neural networks have outperformed traditional methods by such a wide margin on the labeled faces in the wild (LFW) benchmark \cite{huang2007lfw} that this once challenging benchmark is now considered quite easy, at least under the conventional verification protocol.
With these improved representations, face recognition based on DCNNs can now, theoretically, be used in more complicated scenarios, e.g., to identify criminals in surveillance camera images.

However, the identification problem introduces new and different challenges compared to the verification scenario.
While verification requires only a single $1\colon\!1$ comparison, identification requires $1\colon\!N$ comparisons -- of a probe sample with templates from many subjects in a gallery, e.g., the watch-list of criminals.
Depending on the gallery size $N$, finding the correct identity can be a much harder task than simply performing a correct $1\colon\!1$ comparison.
Furthermore, in many scenarios the gallery can combine multiple images of each enrolled subject to build a more effective template.
An even more subtle aspect of the identification problem is that, in many scenarios, most subjects in probe images are not contained in the gallery at all.
Going back to our previous example, we would hope that the majority of the population is not present in a criminal database.
Thus, the recognition algorithm should be able to work under an $M\colon\!N$ open-set protocol, with the ability to detect and ignore probe samples with identities that are not present in the gallery.
This is achieved by giving low similarity values to all known subjects in the gallery, which is a more intricate task than one would expect.
Contrary to some people's intuition, good open-set recognition is not as simple as thresholding a $1\colon\!1$ verification approach on raw similarity scores.
Although this thresholding does reject some unknowns, we show that there are more effective techniques.

Open-set recognition is clearly desirable for many biometric recognition systems, particularly face.
For example, surveillance cameras in airports capture people and compare their faces with a watch-list of \emph{known} criminals.
The airport staff, which is not included in the watch-list but regularly passes through the eye of the camera, should not confuse the algorithm.
Hence, this list of known, but uninteresting people can be seen as \emph{known unknowns} during training.
Finally, many \emph{unknown unknowns}, i.e., passengers that are not on the watch-list and sojourn in the airport need to be ignored by the face recognition algorithm.

In this paper we introduce a small open-set face identification evaluation protocol based on the widely used LFW dataset, which previously has mainly been used for evaluating face verification systems.
Particularly, we introduce \emph{known unknowns}, i.e., probe images at query time with identities that were used during training but are not present in the gallery; and \emph{unknown unknowns}, i.e., subjects at query time whose identities have never been seen by the system, neither during training nor during enrollment.
An illustration of these concepts is shown in \fig{teaser}.
While the LFW dataset is now considered easy under a verification protocol, we show that under our open-set identification protocol for this dataset is still quite challenging.
Note that this is the first protocol that deals with known and unknown unknowns; similar protocols can be generated for other datasets.

We evaluate three different approaches to open-set face recognition.
Each of these algorithms operate on the same high-quality deep features, which we extract using the publicly available VGG face network \cite{parkhi2015deep}.
First, we evaluate a standard $1\colon\!1$ verification-like technique that is applied to deep features \cite{sun2014deep,chen2016unconstrained}, i.e., we compute cosine similarities between deep features in gallery templates and probes; to reject unknowns we use a threshold on the similarity score.
Second, we perform the standard linear discriminant analysis (LDA) technique \cite{zhao98discriminant} on top of the deep features, making use of the known unknowns during training.
We then use the learnt projection matrix to project the original deep feature vectors to the LDA subspace and compare gallery templates to probes via cosine similarity.
Finally, to model probabilities of inclusion with respect to the support of gallery samples, we train an extreme value machine (EVM) \cite{rudd2017extreme} on cosine distances between deep features, again using known unknowns during training.
We show that the raw cosine similarity performs well in a closed set scenario but not in an open-set setting, LDA can detect known unknowns very well but not the unknown unknowns, while EVM can handle both open-set cases with similar precision.

\section{Related Work}
\label{sec:related}

The need for open-set face recognition has been widely acknowledged for well over a decade \cite{grother2003frvt,phillips2011evaluation}.
Nonetheless, only a few works, e.g., \cite{sun2015deepid3,ekenel2009open,stallkamp2007video,li2005open,best-rowden2014unconstrained,liao2014benchmark} have addressed the problem by predominantly focusing on obtaining an ad hoc rejection threshold on similarity score under an open-set evaluation protocol \cite{phillips2011evaluation}.
For example, Best-Rowden \etal{best-rowden2014unconstrained} showed that a simple thresholding of a commercial of the shelf (COTS) algorithm works perfectly for verification, but does not provide decent open-set identification performance.
The development of classifiers that explicitly model probability of inclusion \cite{rudd2017extreme} of probe samples with respect to a region of known support of the gallery has received far less attention in the face recognition community.
For security-oriented applications where the enrollment process must be quick, the cost of false alarms is high, and the cost of missed alarms is even higher, the notion of using an ad hoc rejection threshold on similarity is problematic because the concept of \emph{unknown} may change as more samples are enrolled and data bandwidth is variable, so a \emph{one size fits all} threshold may not work well.
A classifier that can efficiently be retrained with each enrolled gallery template to autonomously assess the probability that probe data comes from regions of known support on behalf of the gallery while considering variable data bandwidths is a far more appealing alternative.

Thus, the motivation for applying classifiers that are open-set-by-design to face recognition problems is manifested.
Several such classifiers have been developed in the computer vision community \cite{rudd2017extreme,scheirer2013toward,scheirer2014probability,bendale2016towards}, but their application has been limited to toy problems on modifications of canonical computer vision datasets like MNIST \cite{lecun1998mnist}, or to generic object recognition problems like the ImageNet challenge \cite{russakovsky2015imagenet}.
However, object recognition problems inherently differ from biometric applications insofar as they are far more coarse-grained, the notion of enrollment does not exist, and deep learning solutions can be obtained by training an end-to-end network on the training set and using that end-to-end network as a classifier.

Face identification systems that use deep features \cite{parkhi2015deep,sun2014deep,taigman2014deepface,chen2016unconstrained}, by contrast, use truncated forward passes over pre-trained networks to extract features at enrollment or query time.
The networks are trained in an end-to-end manner on labeled face identities, which generally differ from the identities enrolled into gallery templates.
Templates are constructed during enrollment, e.g., by collecting extracted feature vectors from several images of each given subject.
At query time, probe templates consisting of one or more extracted feature vectors of one subject, are matched against gallery templates.
The identification procedure commonly takes the form of finding the gallery template with the sample of maximum  similarity to the corresponding probe.
Cosine is a common measure of similarity between feature vectors extracted from a face network \cite{sun2014deep,chen2016unconstrained}.
Particularly, when templates vary in number of images, feature vectors are sometimes aggregated for a given identity prior to matching, e.g., by taking the mean feature vector \cite{parkhi2015deep,chen2016unconstrained}.

To further enhance the performance of a similarity measure between two feature vectors, a common technique is to learn a transformation matrix across the training set that increases the similarity of vectors from the same class, while decreasing the similarity of vectors from different classes.
Linear discriminant analysis (LDA) \cite{bishop2006pattern}, which learns a subspace projection that minimizes the ratio of intra-class to inter-class variance over the training set, is one of the fundamental techniques that has been applied to face recognition \cite{zhao98discriminant} in the past.
More advanced techniques include joint Bayesian \cite{sun2014deep} and triplet-loss \cite{parkhi2015deep,sankaranarayanan2016triplet} embeddings, but they require more training data.

While ad hoc thresholding of raw similarity measures between features and projections thereof can lead to improved open-set face recognition performance, in this paper we compare the results of using such techniques to using a lightweight classifier -- the extreme value machine (EVM) introduced by Rudd \etal{rudd2017extreme} -- that is open-set by design.
EVM uses statistical extreme value theory (EVT) based calibrations over margin distributions to obtain a \emph{probability of sample inclusion} of each probe sample with respect to a gallery template.
In doing so, it implicitly accounts for varying data bandwidths, yielding superior bounds on open space to those of a raw thresholded similarity function.

EVM has some similarities to cohort normalization techniques \cite{tistarelli2015cohort}, parts of it can be viewed as an improved way of zero normalization (Z-norm) \cite{auckenthaler2004score}.
However, while Z-norm assumes Gaussian distribution of the data and takes into account all cohort data points -- even if they are far away from the gallery template -- EVM only considers the points with the highest similarities, and fits a Weibull distribution on half the distance in order to model \emph{margin} distributions.
While Scheirer \etal{scheirer2010robust} showed that EVT can be successfully applied for score normalization in score fusion of biometric algorithms, in this paper we investigate its application to build gallery templates for open-set recognition.

\section{Approach}
\label{sec:approach}
As many readers might not be familiar with open-set evaluation, let us first introduce our exemplary implementation of an open-set protocol and explain the evaluation in more detail, before we discuss the tested algorithms.

\subsection{Open-Set Face Recognition Protocol}
\label{sec:database}

Open-set face recognition has not been studied, in part due to the dearth of open-set evaluation protocols for face databases.
Although the IJB-A dataset \cite{klare2015ijba} provides an open-set protocol, IJB-A has a lot of different issues such as missing annotations, many profile and low quality images, and huge template sizes for both enrollment and querying.
Hence, all these issues have to be solved before researchers can tackle the open-set problem using this dataset.

Also, neither IJB-A nor any other publicly available face recognition dataset provides an evaluation protocol to test open-set identification with both known unknowns and unknown unknowns.
For example, \cite{ekenel2009open,stallkamp2007video} only tests known unknowns, while other protocols \cite{liao2014benchmark,li2005open,best-rowden2014unconstrained} have disjoint training and enrollment set, which only allows to test unknown unknowns.
Thus, we implemented our own evaluation protocol, which is non-random, simple and can easily be implemented.
We chose to generate an open-set protocol for the labeled faces in the wild (LFW) dataset \cite{huang2007lfw}, for several reasons.
First, LFW is publicly available, well-investigated, and contains relatively unconstrained imaging conditions.
Second, LFW is large enough to provide meaningful results, yet it is small enough that experiments can be run using a normal desktop computer.
Finally, LFW contains several identities, for which only a single image is present -- which fits perfectly into our open-set concept.

We have split the identities in the LFW dataset into three groups.
Those 602 identities with more than three images are considered to be the \emph{known} population.
The 1070 identities with two or three images are the \emph{known unknowns}, while the 4096 identities with only one image are considered to be \emph{unknown unknowns}.
The training set $\bf T$ contains the first three images (i.e., the images ending with \texttt{0001.jpg}, \texttt{0002.jpg} and \texttt{0003.jpg}) for each of the known identities, and one image (the image ending with \texttt{0001.jpg}) for the known unknowns.
The enrollment set $\bf G$ is composed of the same three images for each of the known identities, which makes the protocol \emph{biased}.
Note that there are no unknowns (neither known nor unknown unknowns) inside the enrollment set.

Finally, we created four different probe sets, \texttt{C}, \texttt{O}$_1$, \texttt{O}$_2$, and \texttt{O}$_3$.
The closed-set \texttt{C} contains the remaining images $\bf S$ of the known subjects, where the number of probe images per identity can vary between 1 and 527 (i.e., for George W. Bush).
This set is used to evaluate closed-set identification and verification.
Probe set \texttt{O}$_1 = \bf S \cup \bf K$ contains the same images as in the closed probe set \texttt{C}, and additionally the images $\bf K$ from the known unknowns, which were not part of the training set, one or two images per identity.
\texttt{O}$_2=\bf S \cup \bf U$ contains the closed-set images of \texttt{C} and the unknown unknowns $\bf U$, one per identity, which have not been seen during training and enrollment.
Finally, the probe set \texttt{O}$_3$ contains all probe images, including known subjects, known unknowns, and unknown unknowns, i.e., \texttt{O}$_3 = \bf S \cup \bf K \cup \bf U$.

\subsection{Evaluation}
\label{sec:evaluation}
The closed-set evaluation uses standard cumulative match characteristics (CMC) curves and receiver operating characteristic (ROC) curves.
Open-set recognition uses the detection and identification rate (DIR) curves as proposed in the Handbook of Face Recognition \cite{phillips2011evaluation}.

Cumulative match characteristics curves plot the identification rate, a.k.a. the recognition rate, with respect to a given rank.
For each known probe $P\in \bf S$ of identity $p$, the rank $r$ is computed as the number of subjects $g$ in the gallery that are more similar than the correct subject, i.e.:
{\small
\begin{equation}
  \label{eq:rank}
  \mathrm{rank}(P) = \Bigl|\bigl\{ G^g \mid s(G^g,P) \geq s(G^p,P); G^g\in \bf G\bigr\}\Bigr|
\end{equation}
}%
for a given similarity function $s(\cdot,\cdot)$.
This means that rank $r=1$ is assigned when the correct subject is the most similar one.
The CMC curve plots illustrate the relative number of probes that have reached \emph{at least} rank $r$.

Detection and identification rate curves plot the identification rates with respect to the false alarm rates, which should not be confused with false acceptance rates in ROC curves.
For a given similarity threshold $\theta$, a false alarm is issued when the similarity of an unknown probe $P\in \bf K \cup \bf U$ to \emph{any} of the gallery subjects is higher than $\theta$.
The false alarm rate computes the average probability of these \cite{phillips2011evaluation}:
{\small
\begin{equation}
  \label{eq:far}
  \mathrm{FAR}(\theta) = \frac{\Bigl|\bigl\{P \mid \max\limits_g s(G^g,P) \geq \theta;\ P \in \bf K \cup \bf U \bigr\}\Bigr|}{\bigl|\bf K \cup \bf U\bigl|}\,,
\end{equation}
}%
while the detection and identification rate for a given rank $r$ is calculated on the known probe set, and given by \cite{phillips2011evaluation}:
{\small
\begin{equation}
  \label{eq:dir}
  \!\!\mathrm{DIR}(\theta) = \frac{\Bigl|\bigl\{P \mid \mathrm{rank}(P) \geq r \wedge s(G^p,P) \geq \theta; P \in \bf S \bigr\}\Bigr|}{\bigl|\bf S\bigl|}\,.\!\!
\end{equation}
}%
When plotting the DIR curve, different values for the threshold $\theta$ can be computed based on a given false alarm rate $\vartheta$.
After sorting the scores from \eqref{eq:far} descendantly:
{\small
\begin{equation}
  \label{eq:scores}
  \mathrm{scores} = \mathrm{sort}(\{\max\limits_g s(G^g,P) \geq \theta \mid P \in \bf K \cup \bf U \})
\end{equation}
}%
the threshold can be computed by taking the smallest score $\theta > \theta'$, where:
{\small
\begin{equation}
  \label{eq:theta}
  \theta' = \mathrm{scores}\Bigl[\floor[\big]{\vartheta \cdot | \bf K \cup \bf U |}\Bigr]
\end{equation}
}%
Note that the threshold $\theta$ does not exist when $\theta'$ is already the maximum score.

\subsection{Compared Methods}
\label{sec:methods}

\subsubsection{Cosine Similarity}
Most face recognition algorithms that work on deep features simply apply a cosine similarity between pairs of deep feature vectors.
Thus, we obtain a baseline measurement by computing the cosine similarity between the deep feature vectors of gallery template $G^g$ of subject $g$ and probe $P$.
Since each gallery template is composed of three deep feature vectors: $G^g = (G^g_0, G^g_1, G^g_2)$, we apply two strategies: First, we compute three similarities and take the maximum value, which has been shown to provide the best performance in handling several scores \cite{gunther2016frice}:
\begin{equation}
  \label{eq:max-cos}
  s_{\mathrm{max}}(G^g, P) = \max\limits_{i\in\{0,1,2\}} \cos\bigl(G^g_i, P\bigr)\,,
\end{equation}
and second, we average the three deep features vectors \cite{chen2016unconstrained}:
\begin{equation}
  \label{eq:mean}
  \bar{G}^g = \frac13\sum\limits_{i\in\{0,1,2\}} G^g_i
\end{equation}
and compute the similarity between this average and the probe feature vector:
\begin{equation}
  \label{eq:avg-cos}
  s_{\mathrm{avg}}(\bar{G}^g, P) = \cos\bigl(\bar{G}^g, P\bigr)\,.
\end{equation}
Without further processing, this similarity is used inside the evaluation.

\subsubsection{Linear Discriminant Analysis}
To introduce a learning algorithm that can make use of the known unknowns during training, we select linear discriminant analysis (LDA) to learn a projection matrix $W$.
First, we compute a principal component analysis (PCA) projection matrix.
After projecting all training features $T\in\bf T$ into the PCA subspace, we train the LDA with the 603 classes of the training set $\bf T$, i.e., one class for each of the 602 known gallery subject, and one class containing the known unknowns.
For more details on how to train a PCA+LDA projection matrix, please refer to \cite{zhao98discriminant,yambor2000analysis}.
Finally, we project all enrollment and probe features into the combined PCA+LDA subspace using projection matrix $W$:
\begin{align}
  y_{G^g_i} &= W^T G^g_i\,, & y_{\bar G^g} &= W^T \bar G^g\,, & y_{P} &= W^T P\,.
\end{align}
Scores are computed using the functions introduced in \eqref{eq:max-cos} and \eqref{eq:avg-cos} on the projected features:
\begin{align}
  \label{eq:max-lda}
  s_{\mathrm{max}}(y_{G^g_i}, y_{P}) && s_{\mathrm{avg}}(y_{\bar G^g}, y_{P})&\,.
\end{align}

\subsection{Extreme Value Machine}
\label{sec:evm}

For a third approach, we choose the extreme value machine (EVM) introduced by Rudd \etal{rudd2017extreme}.
While EVM was formulated to handle generic classification tasks, we utilize the algorithm to perform biometric identification.
The EVM classifier uses statistical extreme value theory (EVT) \cite{fisher1928limiting} to perform nonlinear, kernel-free classification, optionally in an incremental learning setting.
The classifier fits an EVT distribution per point over several of the nearest fractional radial distances to points from other classes, and uses a statistical rejection model on the resultant cumulative distribution function (CDF) to model probability of sample inclusion (PSI or $\Psi$).
Taking a fixed number of the data point and distribution pairs per class that optimally summarize each class of interest yields a compact probabilistic representation of each class in terms of extreme vectors (EVs).

We tailor EVM to a face identification similarity function by letting each feature vector be associated with an identity.
Deviating slightly from the original formulation, in which the fractional distance over which to fit EVT distributions was assumed to be $\alpha=0.5$ times the distance (cf.~\eqref{eq:dist}) to formalize the classifier in terms of fitting margin distributions,  we formalize the distance multiplier in terms of hyperparameter $\alpha$.
With $\alpha\neq0.5$, EVM no longer models maximum margin distributions, but rather a biased margin distribution.
However, the margin distribution theorem from Rudd \etal{rudd2017extreme}, which governs the functional form of the EVT distribution for modeling probability of sample inclusion $\Psi$, still holds -- dictating that the low tail of multiplied distances will follow a Weibull distribution.
Applying a statistical rejection model to the resultant CDF, each feature vector within the gallery will have its own $\Psi$ distribution.
Denote the $i$th feature vector for gallery subject $g$ as $G^g_i$.
The resultant probability that probe $P$ is associated with $G^g_i$ is given by:
\begin{equation}
  \label{eq:weibull_cdf}
  \Psi(G^g_i,P;\kappa^g_i, \lambda^g_i) = \exp^{-\left(\frac{d(G^g_i,P)}{\lambda^g_i}\right)^{\kappa^g_i}},
\end{equation}
where $d(G^g_i,P) = 1-\cos(G^g_i,P)$ is the cosine distance of a probe $P$ from a subject's feature $G^g_i$, and $\kappa^g_i$, $\lambda^g_i$ are Weibull shape and scale parameters, respectively.
These parameters are obtained for each gallery feature $G^g_i$ by computing all distances:
\begin{equation}
  \label{eq:dist}
  \mathrm{dist} = \bigl\{\alpha \cdot d(G^g_i,T) \mid t \neq g; T \in \bf T\bigr\}
\end{equation}
for all training set features $T\in\bf T$ with identity $t$, which do not correspond to the gallery identity $g$.
A Weibull distribution is fit to the low tail of $\mathrm{dist}$:
\begin{align}
  \label{eq:dist-tau}
  \mathrm{dist}_\tau &= \bigl\{d \mid d \in \mathrm{dist} \wedge d < \theta^\tau\bigr\}\qquad\text{with}\\
  \theta^\tau &= \max_\theta\ \Bigl|\bigl\{d \mid d \in \mathrm{dist} \wedge d \leq \theta \bigr\}\Bigr| = \tau\,,
\end{align}
where the tail size $\tau$ represents a second hyperparameter of EVM.
For details on how to fit Weibull distributions on $\mathrm{dist}_\tau$, please refer to \cite{rudd2017extreme} or the MetaRecognition library.\footnote{The C and Python implementation of \emph{libMR} is provided in:\\{\scriptsize\url{http://pypi.python.org/pypi/libmr}}}

Another modification that we make to the original EVM algorithm is that we retain all EVs rather than select only the most informative ones.
The main purpose of model reduction in \cite{rudd2017extreme} is to maintain compact representations in incremental learning settings.
To be comparable to the other two algorithms, instead we perform scoring in two different ways, one where we compute the maximum probability over each feature $G^g_i$ inside the gallery template $G^g$:
\begin{equation}
  \label{eq:max-evm}
  s_{\mathrm{max}}(G^g, P) = \max\limits_{i\in\{0,1,2\}} \Psi(G^g_i, P, \kappa^g_i,\lambda^g_i)\,.
\end{equation}
For the other technique, we use the average $\bar{G}^g$ from \eqref{eq:mean} for each gallery template, compute distances between $\bar G^g$ and training set features $T$ as in \eqref{eq:dist}, and compute a Weibull fit on the tail of them in order to obtain $\bar\kappa^g$ and $\bar\lambda^g$.
The final probability of inclusion is given as:
\begin{equation}
  \label{eq:avg-evm}
  s_{\mathrm{avg}}(G^g, P) = \Psi(\bar G^g, P, \bar\kappa^g, \bar\lambda^g)\,.
\end{equation}

\section{Experiments}
\label{sec:experiments}

\begin{figure*}[t!]
  \vspace*{-4ex}
  \centering
  \subfloat[Closed-set CMC for different $\alpha$\label{fig:optimization:cmc1}]{\includegraphics[width=.9\columnwidth, page=1]{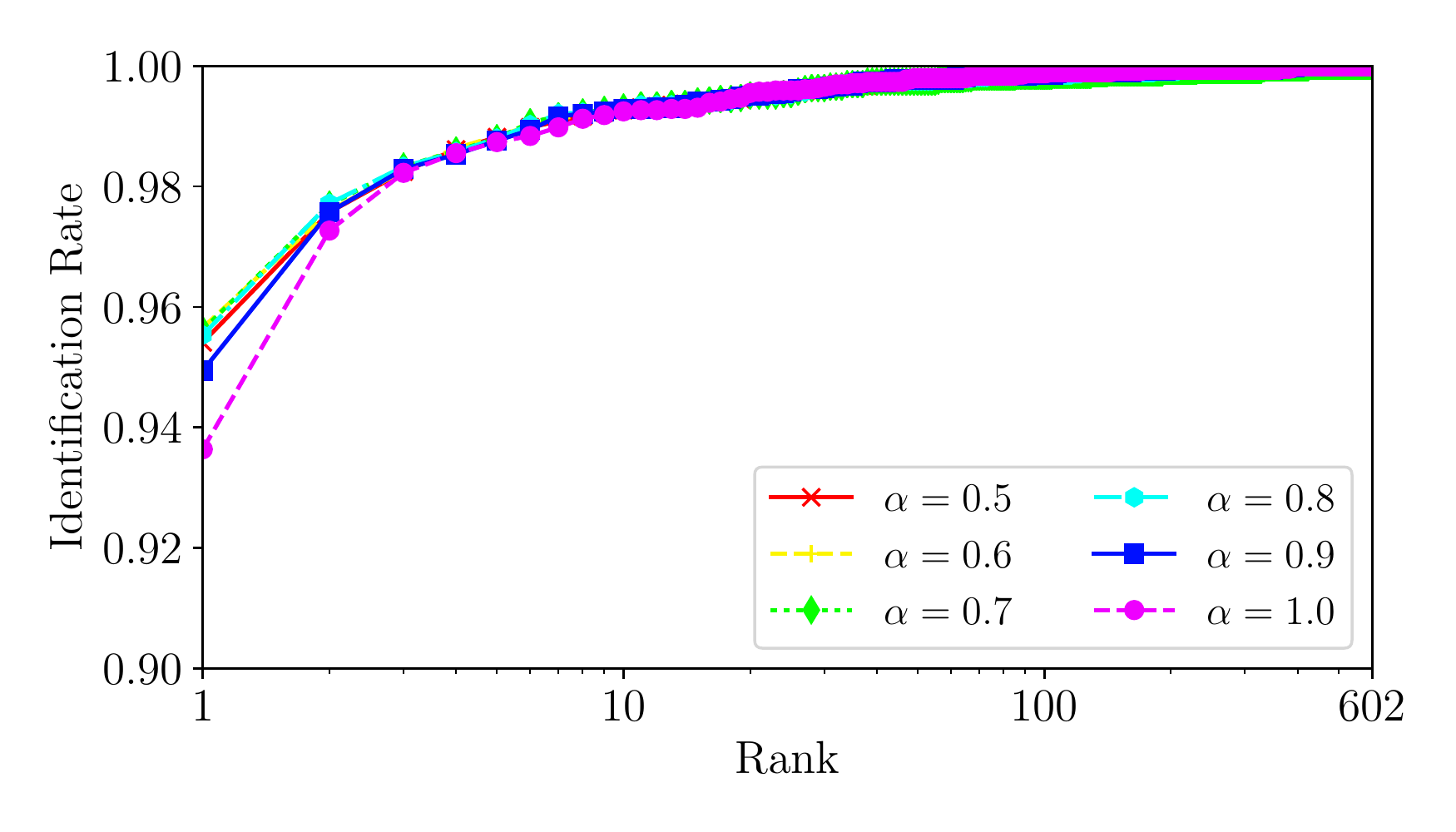}}\qquad%
  \subfloat[Open-set DIR for different $\alpha$\label{fig:optimization:dir1}]{\includegraphics[width=.9\columnwidth, page=3]{EVM_ranges_averaged}}\\[-2ex]
  \subfloat[Closed-set CMC for different $\tau$\label{fig:optimization:cmc2}]{\includegraphics[width=.9\columnwidth, page=1]{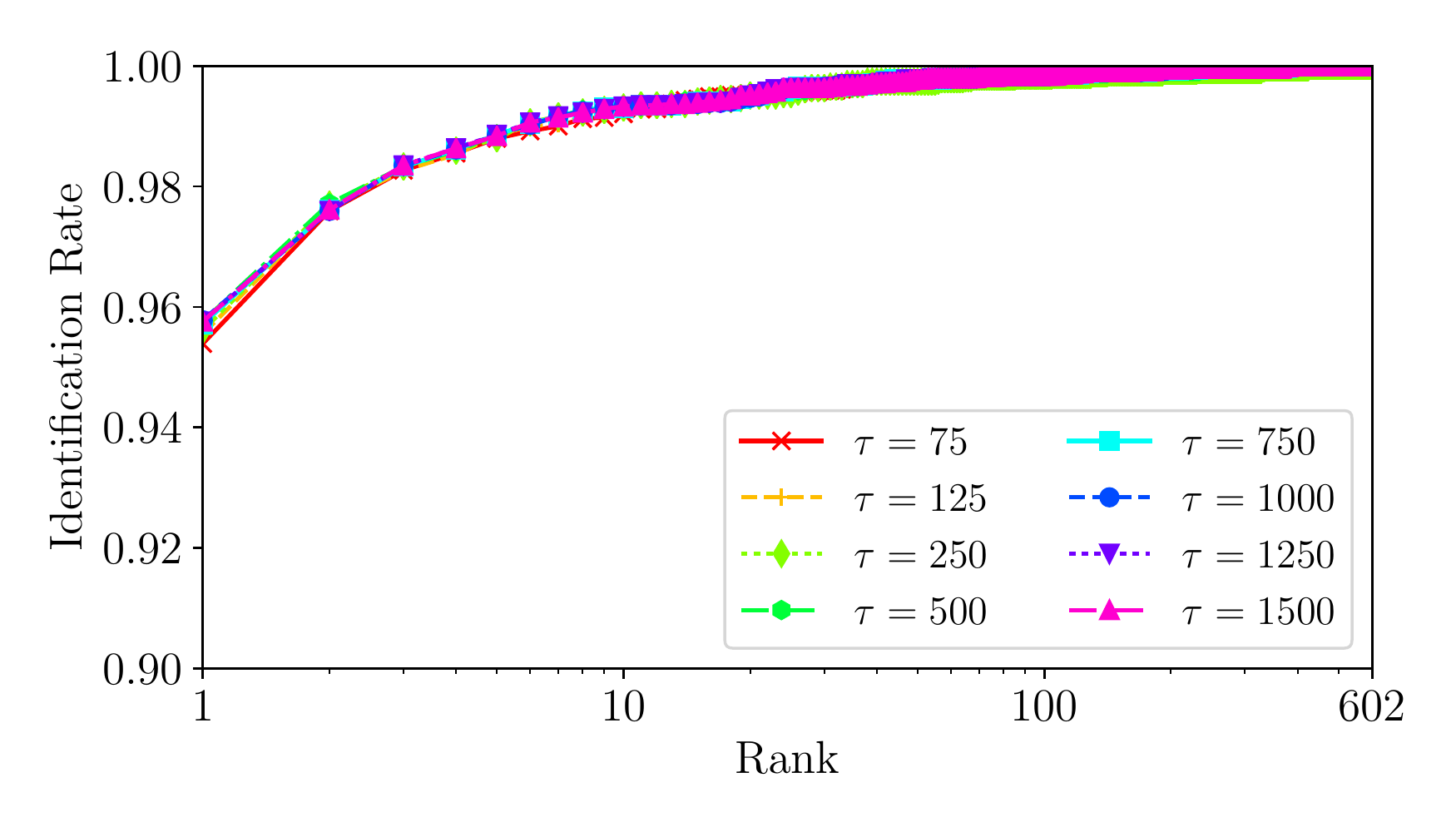}}\qquad%
  \subfloat[Open-set DIR for different $\tau$\label{fig:optimization:dir2}]{\includegraphics[width=.9\columnwidth, page=3]{EVM_tailsizes_averaged}}\\[-1ex]
  \cap{fig:optimization}{EVM Hyperparameter Selection}{Two different parameters of EVM are optimized: the distance multiplier $\alpha$ and the tail size $\tau$. Both are evaluated using the CMC curve on probe set \texttt{C} and the DIR curve on probe set \texttt{O}$_3$.}
\end{figure*}

We conduct experiments on our novel open-set LFW protocol.
For feature extraction, we use the VGG face network.\footnote{The \textit{Caffe} model of the VGG network was downloaded from:\\ {\scriptsize\url{http://www.robots.ox.ac.uk/~vgg/software/vgg_face}}}
We employ the \textit{Bob} signal processing library \cite{anjos2012bob} to align the eye locations\footnote{Annotations are provided under:\\{\scriptsize\url{http://lear.inrialpes.fr/people/guillaumin/data.php}}} of input images to fixed locations $(82,112)$ and $(142,112)$ pixels, to approximately mimic the alignment required for the VGG face network \cite{parkhi2015deep}.
We use \textit{Caffe} \cite{jia2014caffe} to extract the 4096-dimensional \texttt{fc7} layer features from the VGG network, after removing all following layers from the network \texttt{prototxt}, including the \texttt{ReLU} layer.
We perform EVT calibration using \textit{libMR} \cite{scheirer2011metarecognition}.
Finally, to compute and plot the closed- and open-set evaluation results, \textit{bob.measure}\footnote{\textit{Bob} and the \textit{bob.measure} package can be found under:\\{\scriptsize\url{http://www.idiap.ch/software/bob}}} is employed.

\subsection{Hyperparameter Selection}
In the first set of experiments, we evaluate the effects of different hyperparameters for LDA and EVM.
We only test different hyperparameters using the averaging approach, as experiments run faster with it, and we evaluate closed-set identification using probe set \texttt{C}, and open-set recognition using the combined probe set \texttt{O}$_3$.
For LDA, only a single hyperparameter is optimized, which is the number of PCA components retained.
After experimenting with several values, we find that retaining $99\,\%$ of the PCA variance gave the best results, leading to a final PCA+LDA subspace size of 256 dimensions.

EVM has two hyperparameters: the distance multiplier $\alpha$ and the tail size $\tau$.
We start optimizing $\alpha$ by setting the tail size to a hand-picked value of $\tau=250$.
Closed-set and open-set evaluation of different values of $\alpha$ are shown in \sfig{optimization:cmc1} and \subref{fig:optimization:dir1}, respectively.
For closed-set, differences can only be seen in the very low ranks, after rank 5 all CMC curves seem to overlap completely.
The best $\alpha$ value is between $0.6$ and $0.8$ and performance degrades slightly for smaller and larger values of $\alpha$.
Nevertheless, rank 1 identification rates are very high and do not vary substantially with any choice of $\alpha$ in the tested range.
When examining the open-set performance in \sfig{optimization:dir1}, we can see slightly larger differences for different values of $\alpha$.
For large $\alpha$ values, many unknown probes have the probability of 1 to belong to one of the gallery subjects and, hence, thresholds for low false alarm rates cannot be computed.
When the weight multiplier $\alpha$ gets lower, fewer of these cases occur.
Based on both plots, we decide to use $\alpha=0.7$ as a good trade-off between closed-set and open-set performance.

To evaluate different values for the tail size $\tau$, we keep $\alpha=0.7$ fixed.
Examining the closed-set CMC curve in \sfig{optimization:cmc2}, we can again see very little difference.
Generally, larger tail sizes seem to lead to better rank 1 identification rates, but already for rank 3, there is no apparent difference between any of the tested values.
The open-set DIR curve given in \sfig{optimization:dir2} reveals that the open-set performance deteriorates for high and low tail sizes, while $\tau=500$ seems to provide the best overall performance.

\subsection{Comparison between Methods}
\begin{figure*}[t!]
  \vspace*{-4ex}
  \centering
  \subfloat[Closed-set CMC (\texttt{C})\label{fig:comparison:cmc}]{\includegraphics[width=.9\columnwidth, page=1]{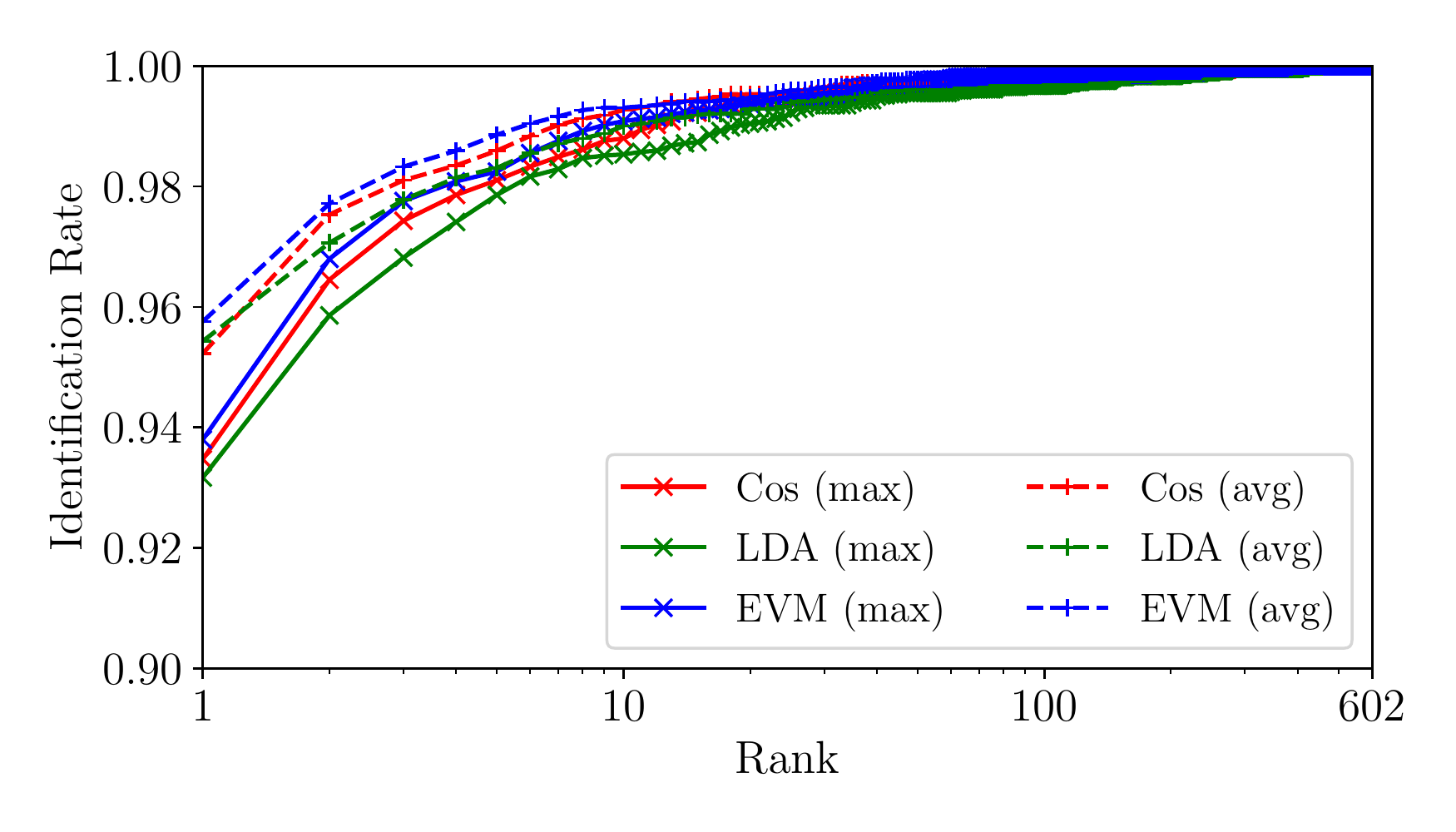}}\qquad%
  \subfloat[Closed-set ROC (\texttt{C})\label{fig:comparison:roc}]{\includegraphics[width=.9\columnwidth, page=2]{VGG_comparison}}\\[-2ex]
  \subfloat[Open-set DIR with Known Unknowns (\texttt{O}$_1$)\label{fig:comparison:dir1}]{\includegraphics[width=.9\columnwidth, page=4]{VGG_comparison}}\qquad%
  \subfloat[Open-set DIR with Unknown Unknowns (\texttt{O}$_2$)\label{fig:comparison:dir2}]{\includegraphics[width=.9\columnwidth, page=5]{VGG_comparison}}\\[-1ex]
  \cap{fig:closed}{Comparison between Methods}{The \subref*{fig:comparison:cmc} CMC curves and the  \subref*{fig:comparison:roc} ROC curves for the closed-set evaluation on probe set \texttt{C}, as well as the open-set DIR curves for \subref*{fig:comparison:dir1} probe set \texttt{O}$_1$ and \subref*{fig:comparison:dir2} probe set \texttt{O}$_2$ are given for all six evaluated methods.}
\end{figure*}

After obtaining the optimal hyperparameters for LDA and EVM, we compare the performances of all three methods, and also with both scoring approaches, i.e., $s_\mathrm{max}$ and $s_\mathrm{avg}$.
The closed-set performance of Cos, LDA and EVM on probe set \texttt{C} is given in \sfig{comparison:cmc} and \subref{fig:comparison:roc}.
As expected and reported \cite{best-rowden2014unconstrained}, both closed-set identification and verification reach very high accuracies, e.g., a rank 1 identification rate of up to 96\,\%.
One apparent observation is that the averaging strategy works better for all of the algorithms, which conforms with prior work \cite{chen2016unconstrained} using deep features for face recognition.
Interestingly, all algorithms perform almost similar in the closed-set identification task shown in \sfig{comparison:cmc}, while EVM has a slight advantage over Cos, and LDA performs worst.
However, when evaluating verification performance in \sfig{comparison:roc}, the simple cosine distance seems not to be as good as either LDA or EVM, indicating that distances are distributed differently for different identities (consisting with \cite{doddington1998sheep}), but both EVM and LDA effectively normalize them out.

More interestingly, looking at the open-set performance in \sfig{comparison:dir1}, where we evaluate the known unknowns of probe set \texttt{O}$_1$, it is obvious that Cos has lost much of its performance, especially in lower FAR areas.
On the other hand, LDA and  EVM have similarly good detection and identification rates.
Here, both algorithms can make use of information about the identities during training.
For example, LDA computes its projection matrix so that all known unknowns are clustered together, and are far from any known subject.
In opposition, EVM does not cluster the known unknowns, but only uses distances to them during training.

And precisely because EVM does not model the known unknowns, it is able to maintain its high performance when confronted with unknown unknowns from probe set \texttt{O}$_2$.
On the other hand, LDA's performance dropped dramatically in \sfig{comparison:dir2} for low FAR values.
We assume that the unknown unknowns do not cluster well in the LDA subspace and are, hence, more similar to gallery subjects.
For larger FAR values, however, LDA still outperformed EVM.
We attribute this to the fact that LDA works well for a biased protocol \cite{lui2012preliminary}, i.e., where identities in the training and test sets are shared.
We assume that EVM, in opposition, is not favored by a biased protocol -- as EVM does not use identity information of other subjects, but treats all distances to other subjects' features identically.

Another interesting point is that the discrepancy between the two modeling approaches, i.e., computing the maximum probability over three points and averaging the model features has an influence on the performance of EVM.
While both in the closed-set evaluations and in the open-set evaluation with known unknowns, the average approach works better, it is the opposite in the open-set evaluations with unknown unknowns.
It seems that with identities not seen during training, having more complex models results in a higher robustness with respect to rejecting unknown unknowns.

\section{Discussion}
\label{sec:discussion}
Due to the fact that the number of unknown probe files is relatively low: $|{\bf K}| = 1334$ and $|{\bf U}| = 4069$, computing low false alarm rates, i.e, $\mathrm{FAR} < 0.001$ was often not possible, and results in that range might not be statistically meaningful.
Hence, the advantage of EVM over LDA in \sfig{comparison:dir2} might not be as significant as it looks.
However, the advantage of EVM over raw cosine distances is obvious, both in the closed set ROC curve in \sfig{comparison:roc}, as well as in the open-set evaluation in \sfig{comparison:dir2} since EVM performs better for almost any FAR value.

It is well-known \cite{lui2012preliminary} that LDA-based face recognition algorithms are highly favored by biased protocols like the one that we have introduced.
This is due to the fact that LDA can make use of class information by clustering these classes together during training.
For unbiased protocols G\"unther \etal{guenther2017face} have found that LDA does not improve over simple distance computations in PCA subspace.
Hence, LDA results on more realistic, i.e., unbiased datasets will most probably be lower.

In opposition, EVM does not make use of the classes during training.
For each feature vector in a gallery template, only distances to all other subjects' feature vectors are computed to model the probability of inclusion.
Theoretically, there is little difference whether these features belong to known or unknown subjects.
Hence, we assume that it does not matter whether to query with subjects seen during training, or with unknown subjects, but we leave the verification of this assumption to future work.
Based on this assumption, we can claim that EVM can handle unknown images better than LDA, and clearly better than using a simple thresholded cosine distance.
Hence, we conclude that open-set recognition is better handled by modeling probability of inclusion with respect to gallery support and then thresholding on the posterior probability estimate, as opposed to thresholding raw similarities.
Note that EVMs are also not limited to using simple distance functions between raw features.
For example, a combination of LDA and EVM -- first projecting the features into the LDA subspace and then applying an EVM in projected feature space -- could be a viable approach.

Due to the relatively small size of the dataset, we did not split it up further into validation and test sets with mutually exclusive subjects.
This is why we illustrated the performance of several hyperparameter choices on the test set, rather than use a subset of non-test data to select one set of hyperparameters.
However, performance differed surprisingly little across hyperparameter choices (cf. \fig{optimization}), and choosing other parameters would not change our results.


\section{Conclusion}
\label{sec:conclusion}

In this paper we have shown that open-set face recognition is a difficult problem, and that simply thresholding similarity scores is a weak solution.
We have experimented with two approaches that are often applied for face recognition: computing cosine distances on deep features, and applying linear discriminant analysis (LDA).
Due to the biased nature of our evaluation protocol, LDA worked favorably over cosine in the open-set evaluations, but still performed poorly when tested with unknown unknowns.
Hence, while LDA might be a proper choice for application when mainly known unknowns occur, in public areas (e.g. in airports) with a high amount of passenger traffic, LDA will not be sufficient.
Interestingly, LDA performed worst in the closed-set identification task, yet performed best in the verification task.

In order to model probabilities of inclusion with respect to gallery templates, we invoked the extreme value machine (EVM).
Without making use of identity information during training, EVM was able to perform well in all of our tests, i.e., closed-set identification, verification and open-set identification.
In all cases, EVM was able to beat the simple cosine distance, which demonstrates that modeling inclusion probabilities improves both closed and open-set identification as well as verification.
Further, we assume that in an unbiased dataset, where training and test sets contain different identities, LDA will perform poorly while EVM will approximately maintain its performance.
How well EVM performs with respect to other score normalization techniques such as Z-norm is left for future work.

Anyways, at a false alarm rate of 0.01 (meaning that 1 out of 100 unknown subjects are assigned to one subject in the gallery) only around 60\,\% of the gallery subjects were correctly identified by EVM.
Revisiting our example in \sec{introduction}, a surveillance system in an airport that captures 100 persons per minute and queries each against a criminal database will have one false alarm per minute -- which usually requires human interaction to resolve -- while failing to identify 40\,\% of the criminals.
Though this is a simplified example, it illustrates that open-set face identification is far from being solved, and additional research is required for real-time surveillance applications.
While the open-set face identification protocol that we have introduced in this paper is a good start, research in the open-set identification space would benefit from larger databases that can be split into training, validation and testing sets, yet contain sufficiently many unknown unknowns to be able to calculate meaningful detection and identification rates at reasonable false alarm rates.

\ifcvprfinal
\section*{Acknowledgment}
This research is based upon work supported in part by NSF IIS-1320956 and in part by the Office of the Director of National Intelligence (ODNI), Intelligence Advanced Research Projects Activity (IARPA), via IARPA R\&D Contract No. 2014-14071600012. The views and conclusions contained herein are those of the authors and should not be interpreted as necessarily representing the official policies or endorsements, either expressed or implied, of the ODNI, IARPA, or the U.S. Government. The U.S. Government is authorized to reproduce and distribute reprints for Governmental purposes notwithstanding any copyright annotation thereon.
\fi

{\small
\bibliographystyle{ieee}
\bibliography{paper}
}

\end{document}